\documentclass[10pt,twocolumn,letterpaper]{article}

\usepackage[pagenumbers]{wacv} 

\definecolor{wacvblue}{rgb}{0.21,0.49,0.74}
\usepackage[pagebackref,breaklinks,colorlinks,allcolors=wacvblue]{hyperref}

\usepackage{tipa}
\usepackage{wrapfig}
\usepackage{array}

\usepackage[accsupp]{axessibility}

\def\wacvPaperID{481} 
\def\confName{WACV}
\def\confYear{2027}

\title{Rohbau3D-Semantics: Dense Semantic and Instance Labels for Point Clouds of Indoor Shell Construction Scenes}

\author{
Lukas Rauch \ \ \qquad \ \ Thomas Braml \\
\\ 
University of the Bundeswehr Munich, Germany\\
{\tt \small \{lukas.rauch, thomas.braml\}@unibw.de}
}

\begin{document}
\maketitle
\begin{abstract}
    We present the Rohbau3D-Semantics dataset, which adds dense semantic and instance-level labels to laser scans of the previously released Rohbau3D domain data from shell construction sites. The dataset comprises 504 point cloud scenes from 14 survey sites, each captured with a terrestrial laser scanner and stored in a scanner-centered local coordinate system. 
    Unlike point clouds sampled from reconstructed meshes or synthetic building models, the data preserve the characteristic properties of real laser scanning, including non-uniform sampling, measurement artifacts, reflections, and fine geometric detail. Combined with manual labels, expert oversight during pointwise annotation, and full-scene review, this yields a high-quality benchmark for semantic scene understanding in complex built environments.
    The dataset includes 17 class labels of building components and one background class, along with instance annotations for object-level analysis. In addition to the 3D point clouds, the release also provides scripts for deriving geometrically aligned panoramic and cube-map representations of color, LiDAR reflection intensity, depth, surface normals, semantic labels, and instance labels from the 3D arrays.
    We describe the acquisition pipeline, annotation protocol, quality-control process, and panorama-generation procedure, and provide benchmark splits and baseline experiments for semantic segmentation. By combining precise terrestrial laser scans, dense annotations, and a wide variety of shell-construction scenes, Rohbau3D-Semantics bridges the gap among synthetic indoor datasets, mesh-derived benchmarks, and smaller construction-specific resources. 
    The dataset is intended to support research on 3D scene understanding, construction-site analytics, and multimodal scene analysis in the built environment.
    
    Access and download the full dataset via \href{https://github.com/RauchLukas/rohbau3d}{github.com/RauchLukas/rohbau3d}.
\end{abstract}
    
\section{Introduction}
\label{sec:intro}

\begin{figure*}[t]
  \centering
  \includegraphics[width=0.93\textwidth]{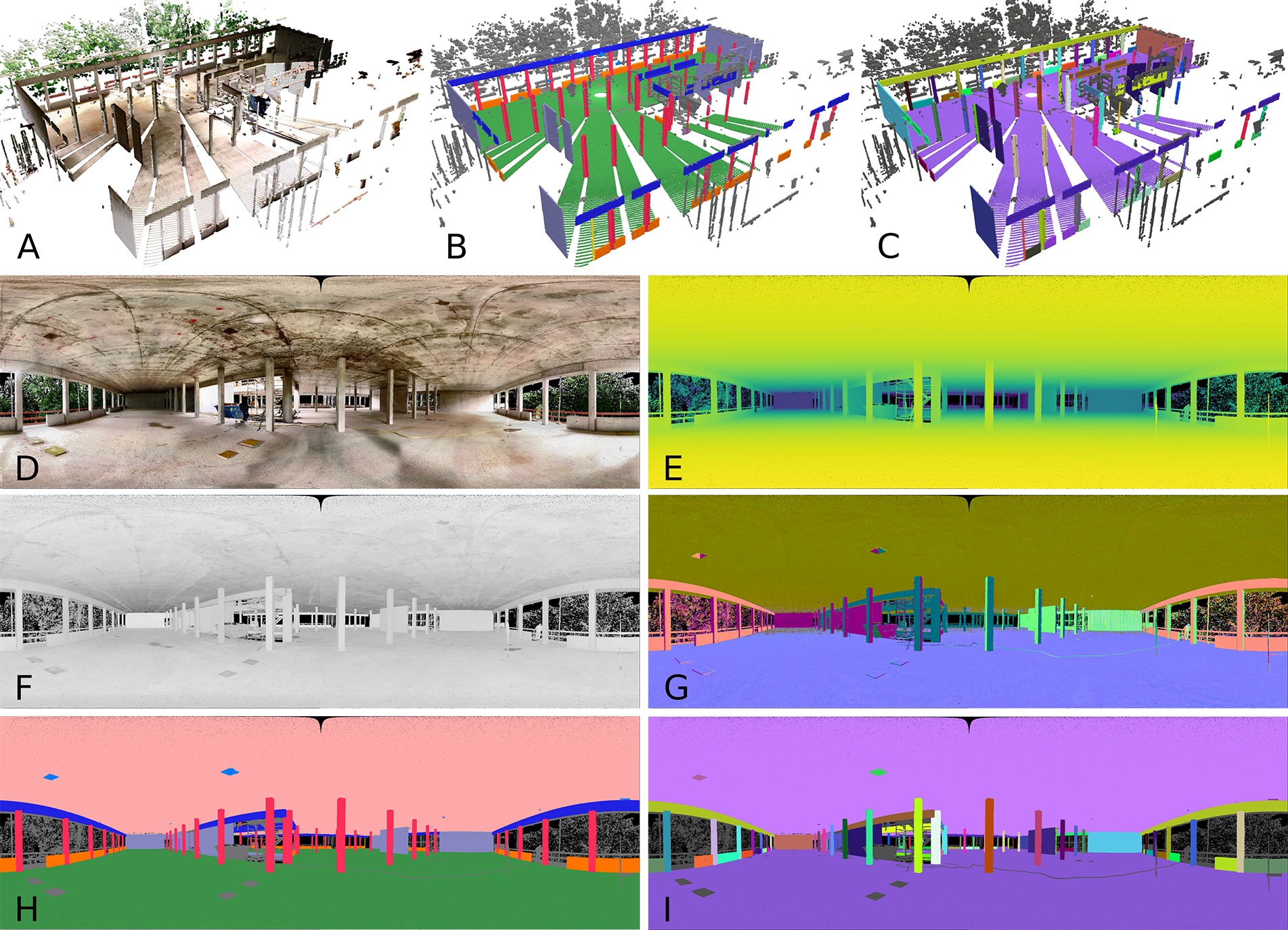}
  \caption{Rohbau3D-Semantics: Dataset features in both 3D and 2D-projected form. A) color, B) semantic classes, C) class instances, D) color, E) depth, F) reflection intensity, G) surface normals, H) semantic classes, I) class instances.}
  \label{fig:teaser}
\end{figure*}

Spatial scene understanding increasingly depends on benchmark datasets and verified validation data that capture complex environments in faithful digital representations. In recent years, major progress in 3D scene understanding has been driven by large datasets of indoor spaces. 
However, the individual works primarily focus on furnished residential and office interiors, using synthetic, CAD-derived scenes \cite{zheng_Structured3D_2020}, RGB-D reconstructions \cite{dai.ScanNet.2017, armeni.Joint.2017}, or high-fidelity laser-scan reconstructions with aligned imagery \cite{chandanyeshwanth.ScanNet.2023}.
Although these benchmarks have substantially advanced semantic segmentation and related tasks in general, they do not address the actual challenges associated with raw data from terrestrial laser scanners and LiDAR measurements of structural-shell conditions.

This limitation is particularly relevant to investigations of shell construction sites. Such scenes are characterized by dynamically changing conditions, temporary structures, stored materials, construction debris, and inconsistent lighting. Fundamental building elements are often unfinished, not yet installed, partially occluded by supporting structures, or concealed by protective coverings. This setting is captured by the German term \textit{Rohbau} \textipa{["ro:baU]}, which denotes the stage of a building project in which only the essential structural shell has been completed. 
Publicly available datasets for shell construction environments remain scarce, despite the importance of this construction phase for digital documentation and progress monitoring \cite{tsige_Automated_2025}, building information modeling (BIM) \cite{tang_BIM_2022, chen_Deep_2019}, and quality assurance for building components \cite{zhang_Quality_2023}. Resources with dense point-wise semantic and instance annotations for an engineering-based understanding of the framework structure are rarer still \cite{rauch.Semantic.2023}. Such annotations are particularly valuable in construction settings because many downstream tasks require not only recognizing object categories, but also distinguishing individual elements. 
While semantic labels support the identification of structurally relevant classes, instance-level annotations are essential for object-wise analysis, such as tracking the installation state of individual components \cite{zou_Method_2025}, 
comparing as-built conditions against BIM models \cite{tsige_Automated_2025, macher_Point_2015}, 
localizing defects, or reasoning about navigable and obstructed space \cite{feng_Integration_2026}.

To address this gap, we present Rohbau3D-Semantics, an extension of the previously released Rohbau3D dataset \cite{rauch_Rohbau3D_2025}. The original Rohbau3D release provided terrestrial laser scans and associated metadata from diverse shell construction sites, but no semantic annotations. Rohbau3D-Semantics extends the raw data with point-wise labels, aligned 2D panoramic projections, benchmark splits, and baseline results for supervised 3D semantic segmentation. The instance labels and linked 2D representations are released to support future instance-level and multimodal research, but these tasks are not evaluated in the present experiments. Although rooted in the construction domain, the dataset may also be relevant to broader work in automation and robotics involving cluttered or structurally degraded built environments \cite{le_3D_2025, pan_Postdisaster_2025}.

\section{Related Datasets}
\label{sec:related_data}

\begin{table*}[!t]
  \caption{Comparison of representative indoor and construction datasets. Domain distinguishes indoor scenes (Interior), building information models (BIM), and construction sites (Constr.). Extent gives the number of sites, areas, rooms, scenes, or scans; Reported scale uses each dataset's native metric. Our reported area is the sum of the floor-plan footprints reconstructed independently on each scan. Overlapping regions observed from multiple scanner positions are counted repeatedly; the value measures cumulative observations rather than unique site floor area.
  Sem./Inst. indicates semantic/instance annotation availability. Modalities: P = point cloud, M = mesh, RGB = color, D = depth, I = intensity, N = normals, and IMU = inertial measurements. Type: R = real and S = synthetic data. ``--'' indicates that the corresponding annotation is unavailable.}
  \label{tab:dataset_comparison}
  \centering
  \footnotesize
  \setlength{\tabcolsep}{2.4pt}
  \renewcommand{\arraystretch}{1.08}
  \begin{tabular}{llccccclc}
    \hline
    Dataset & Domain & Extent & Reported scale & Sensors & Classes & Sem./Inst. & Modalities & R/S \\
    \hline
    S3DIS \cite{armeni.3D.2016} & Interior & 6 areas / 271 rooms & $6{,}000$\,m$^2$ / 215M pts & RGB-D & 13 & Y/Y & P, RGB, D, N & R \\
    ScanNet v2 \cite{dai.ScanNet.2017} & Interior & 1,513 scans & 2.5M frames & RGB-D & 20 & Y/Y & M, RGB, D & R \\
    ScanNet++ v2 \cite{scannetpp.Changelog.2024} & Interior & 1006 scenes & 1.1M DSLR+10M iPhone frames & TLS+DSLR+RGB-D & 3.4k & Y/Y & P, M, RGB, D & R \\
    LiDAR-Net \cite{guo_LiDARNet_2024} & Interior & 14 areas / 465 rooms & $30{,}000$\,m$^2$ / 3.6B points & Mobile LiDAR & 24 & Y/Y & P, RGB & R \\
    VASAD \cite{langlois.VASAD.2022} & BIM & 6 models & 62,000\,m$^2$ & Sim. LiDAR & 11 & Y/N & P, M & S \\
    Hilti-Oxford \cite{lintongzhang.HiltiOxford.2022} & Constr. & 2 sites / 16 sequences & 3,000\,m$^2$ & LiDAR+cam. & -- & N/N & P, RGB, IMU & R \\
    ConSLAM \cite{m.trzeciak_ConSLAM_2022} & Constr. & 1 site / 5 sequences & 1.40\,km trajectories & LiDAR+cam. & -- & N/N & P, RGB, IMU & R \\
    ConSite \cite{yue.ConSite.2026} & Constr. & 3 projects / 55 rooms & 2,641\,m$^2$ / 45.3M pts & TLS & 13 & Y/N & P, RGB & R \\
    \textbf{Ours} & \textbf{Constr.} & \textbf{14 sites / 504 scans} & {\bfseries 66{,}000\,m$^2$ / 12.5B pts} & \textbf{TLS} & \textbf{17+1} & \textbf{Y/Y} & \textbf{P, RGB, D, I, N} & \textbf{R} \\
    \hline
  \end{tabular}
\end{table*}

While several datasets support indoor 3D scene understanding, none captures the same technical scope of shell construction sites.
ScanNet v2 \cite{dai.ScanNet.2017} and S3DIS \cite{armeni.3D.2016} are central real-world indoor benchmarks based on RGB-D reconstructions or registered point samples. ScanNet++ \cite{chandanyeshwanth.ScanNet.2023} adds high-end laser scans, DSLR images, and iPhone RGB-D sequences. These datasets, however, primarily depict completed, furnished interiors rather than scanner-centered raw terrestrial scans of unfinished structural shells.

Structured3D \cite{zheng_Structured3D_2020} provides virtual indoor scenes rendered from CAD models, while VASAD \cite{langlois.VASAD.2022} samples synthetic point clouds from digital building models. Although useful for scalable training, their point distributions arise from rendering or sampling rather than physical sensing. LiDAR-Net \cite{guo_LiDARNet_2024} instead supplies real, unfiltered mobile-LiDAR data with uneven density, noise, and registration errors, but its taxonomy mainly describes common architectural elements and furnishings rather than components specific to structural shells.

Public construction-site data remains scarce. Hilti-Oxford \cite{lintongzhang.HiltiOxford.2022} and ConSLAM \cite{m.trzeciak_ConSLAM_2022} combine cameras, LiDAR, and inertial sensing in construction environments, but target SLAM, registration, or progress monitoring and provide no semantic annotations \cite{sun_Nothing_2023}. ConSite \cite{yue.ConSite.2026} has recently provided TLS data with semantic labels for 13 classes of structural and temporal components, though without instance annotations and only on a small scale.

Table \ref{tab:dataset_comparison} makes the different dataset roles explicit: mobile construction-site datasets target localization and mapping without semantic or instance labels, whereas ConSite provides construction-specific semantic labels but no instances. Against this background, Rohbau3D-Semantics combines real construction-site data with point-wise semantic and instance annotations and linked 2D representations. Among the compared site-related datasets with explicitly reported annotated point totals, it provides the largest collection, with 504 standalone scans and 12.5 billion points.

\section{Dataset Description}
\label{dataset_overview}

Rohbau3D-Semantics builds upon the previously published Rohbau3D dataset, in which the underlying terrestrial laser scans were released as point clouds under a CC BY 4.0 license~\cite{rauch_Rohbau3D_2025}. The present work extends these data with dense point-wise semantic and instance annotations and a reproducible pipeline for deriving corresponding panoramic and cube-map representations. The original publication documents the data acquisition and preprocessing in detail. For completeness, this section summarizes the relevant processing steps before describing the new annotation workflow and the resulting dataset.

\subsection{Data Collection and Processing}

The source data were collected over a 1-year period at multiple construction sites in the Munich area, Germany. 
The survey sites were selected in collaboration with four construction companies and the Munich State Building Authority to ensure diversity in building typologies, construction conditions, and the occurrence of objects.
As a result, the dataset covers a broad range of unfinished indoor environments, with varying surface materials, architectural features, and stages of construction progress visible.

\begin{figure}[!t]
    \centering
    \includegraphics[width=1.0\linewidth]{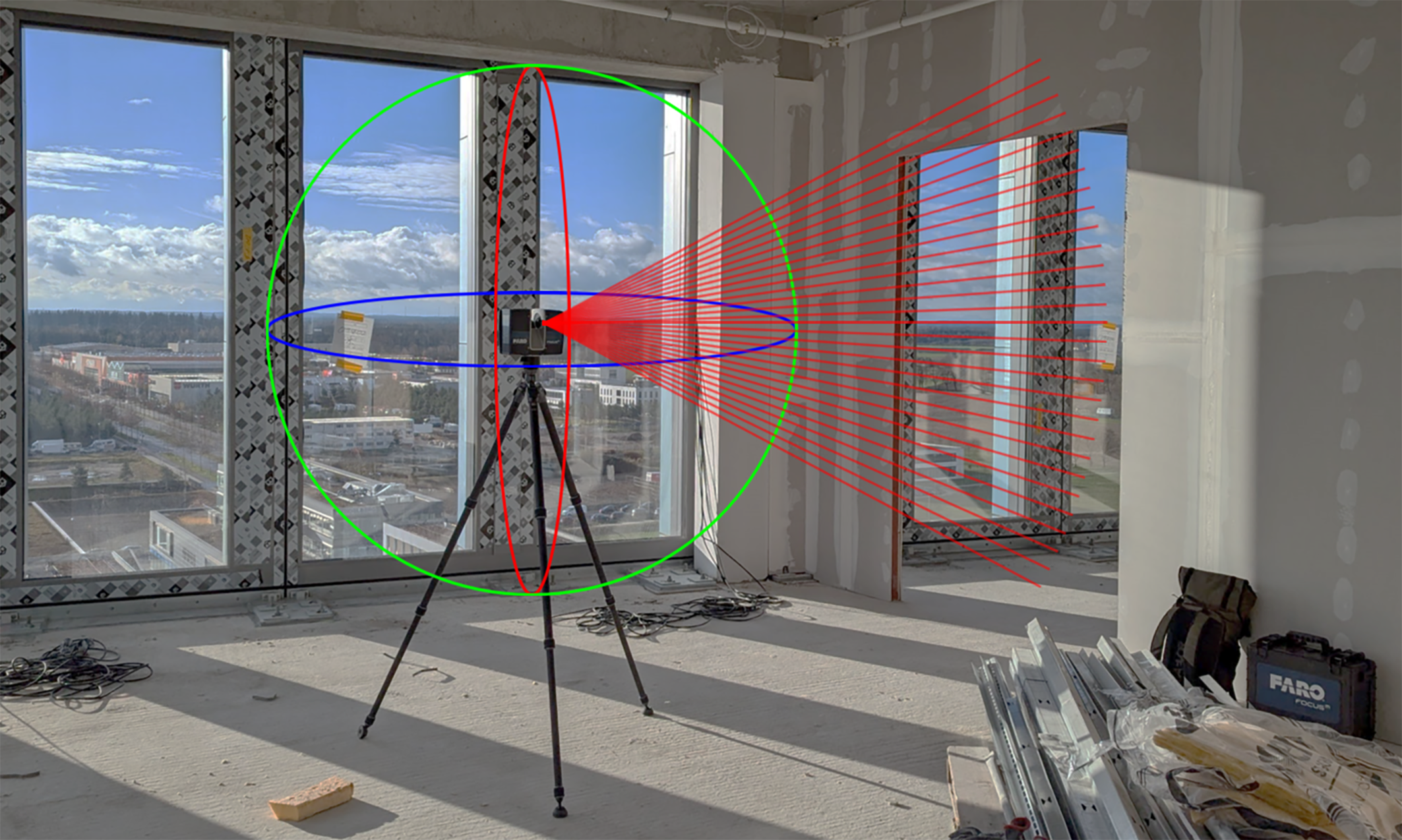}
    \caption{Illustration of the terrestrial laser scanner device that was used to record the data, and a representative construction site scene in the background. The laser sensor head rotates simultaneously around two axes (yaw and roll), thereby mapping the space in a spherical grid. }
    \label{fig:scanner_setup}
\end{figure}

All point clouds were acquired using a steady FARO Focus M70 terrestrial laser scanner shown in Figure \ref{fig:scanner_setup}. This device can capture up to 1 million spherical points per second using LiDAR technology and produce colored point clouds by superimposing spherical high-dynamic-range (HDR) images in post-processing. Construction management and supervision use comparable business-grade laser scanners to measure, document, and evaluate ongoing construction processes. The laser scanner is tripod-mounted, which improves measurement quality compared to handheld devices.

\begin{figure}[!t]
    \centering
    \includegraphics[width=1.0\linewidth]{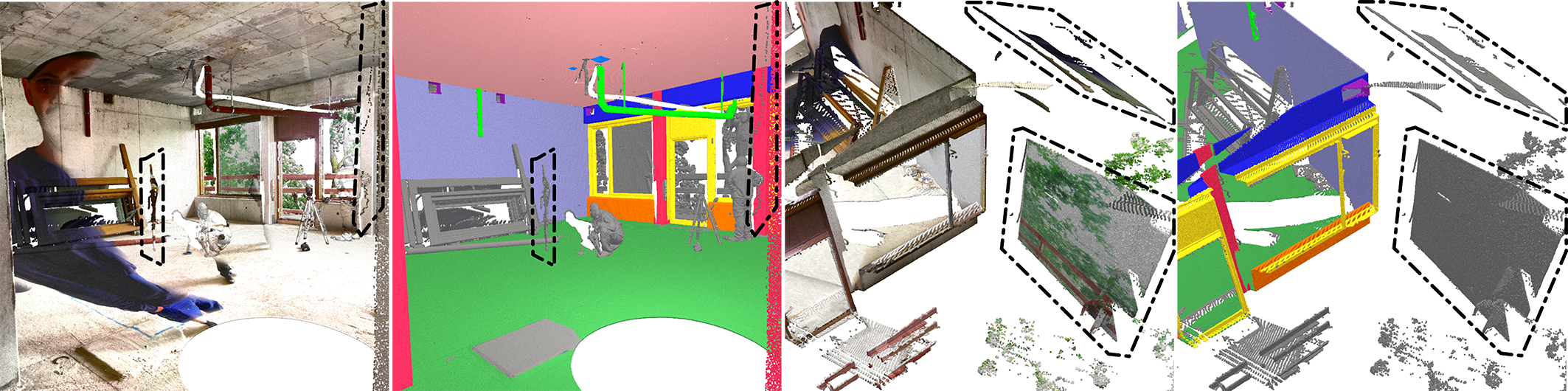}
    \caption{Illustration of reflection noise and ghost points from LiDAR sensing. Reflections typically exhibit nonsensical color textures in RGB space. Ghost points typically appear suspended in the air or near sharp edges. Black polygons highlight such regions that are assigned to the None class (gray).}
    \label{fig:reflections}
\end{figure}

This setup yields high point density, high geometric accuracy, and consistent acquisition results, while preserving characteristic sensing imperfections, such as occlusions, reflections, and measurement artifacts, as illustrated in Figure \ref{fig:reflections}. Scan positions were selected individually on site by a trained civil engineer to mimic the process of creating as-built documentation. 
The individual scans were not registered into a common global coordinate frame. Instead, each scan is stored in its local scanner-centered coordinate system, with the sensor origin located at $x,y,z = (0,0,0)$ in the world coordinate system. 
This representation ensures a consistent, standardized structure across all samples, and a frame-like field of view, comparable to that of mobile LiDAR devices.

Preprocessing was intentionally kept minimal to preserve the raw measurement characteristics. The only geometric filtering step removed points located outside a sphere of radius 25\,m centered at the scanner origin to suppress unstable long-range measurements. No resampling, mesh reconstruction, or geometric regularization was applied. Consequently, the dataset retains the irregular sampling pattern and sensor-specific artifacts of LiDAR sensing, which pose a key challenge for 3D data analysis of actual sensor data \cite{rauch_Rohbau3D_2025}.
Site descriptions and file-format details are provided in the supplementary material; scene-wise feature compendia are available in the dataset repository~\cite{rauch_Rohbau3D_2026}.

\subsection{Data Annotation}

The proposed Rohbau3D-Semantics dataset uses a semantic taxonomy of 17 foreground classes and one background class. The class set was defined as a condensed selection of object categories that are both relevant to building information modeling (BIM) and exist frequently enough in the data to be represented as independent classes. This design avoids overly fine-grained labels for rare objects while taking into account the important components for modeling the building.
The list of annotation classes defines:  (0) None, (01) Ceiling,  (02) Slanted Ceiling, (03)  Ceiling Cutout, (04) Floor, (05) Wall,  (06) Drywall, (07) Wall Cutout, (08) Parapet, (09) Beam, (10) Column, (11) Staircase, (12) Window Cutout, (13) Window, (14) Door Rough Opening, (15) Facade Element, (16) Elevator Doors, and (17) TGA (technical building equipment), often referred to as mechanical electrical plumbing (MEP). Based on the functional unity and geometric continuity, individual objects are further clustered into distinct instances within each semantic segment.

Annotations were created with a professional subcontractor specialized in 3D data labeling. Before large-scale annotation began, the authors trained the provider using detailed guidelines with class definitions, visual examples, and pilot scenes. A condensed version of the finalized annotation guidelines is provided within the supplementary material. An initial validation round on randomly selected samples was then used to identify errors and refine the procedure. Throughout the process, the authors supervised the annotation workflow, and ambiguous cases were resolved in collaboration with a construction-domain expert to ensure consistent labeling.
Point-wise semantic ground-truth labels were produced through interactive point-level selection and iterative refinement. Unlike segment-based workflows in which precomputed clusters form the final labeling units \cite{armeni.Joint.2017, dai.ScanNet.2017}, component boundaries were checked and adjusted directly at the point level.
Class-wise instance-object segmentations were performed on this dataset using iterative polygon-based trimming in the 3D viewer. Point subsets were interactively selected and repeatedly refined from multiple viewing directions, enabling components to be separated in true 3D space with actual 3D geometric knowledge and reducing misinterpretations caused by occlusions or scan artifacts that may appear plausible from a single frontal view but become identifiable as noise from lateral perspectives. This is especially important near reflective surfaces and sharp edges, where LiDAR artifacts can appear plausible in 2D projections but reveal geometric deviations in the 3D representation.

Point-wise instance annotations were defined primarily by geometric continuity rather than by semantic class membership alone. In general, spatially separated elements of the same semantic class were labeled as distinct instances, for example, separate walls within the same room or adjacent windows separated by a narrow wall segment. Occlusion, limited visibility, or local scan artifacts caused an instance split only when the common physical connection remained unclear; disconnected observations were merged when their functional or geometric unity could be inferred from the scene context. Conversely, touching or intersecting elements of the same class were annotated as separate instances when they represented distinct structural components, such as walls meeting at a corner or beams intersecting at a joint. Cutouts and rough openings were treated as separate instances whenever they were spatially distinguishable. Technical building equipment was organized into local instances of related service elements, with close-by elements grouped when their individual boundaries were ambiguous. A special case is the background class None, for which each scene contains exactly one aggregated instance collecting all unassignable points, including clutter, artifacts, and ambiguous structures.

After completion of the annotation process, all 504 scenes were reviewed again by the authors, corrected where necessary, and approved for release. Finally, automated scripts were used to verify annotation completeness and consistency, ensuring that no points remained unlabeled and that each scene contained exactly one background instance of class 0 (None) with instance ID 0.
The external annotation effort amounted to 3,818 hours. Supervision, review, and curation required an additional workload of roughly 10 \% of the actual annotation time.
The complete annotation and quality-control workflow is summarized in the supplementary material.

\subsection{2D Representations}

In addition to the native 3D representation, the release provides scripts for deriving 2D panoramic and cube-map representations from each scene. These views support multimodal use cases and direct image-based inspection of the point-wise features and annotations. The panoramas use an equirectangular projection from the scanner-centered coordinate system onto a spherical image plane. Since each scan's local origin coincides with the scanner position, the projection corresponds directly to the physical acquisition geometry.

Each 3D scan-point is mapped to spherical image coordinates using its \textit{azimuth} and \textit{elevation} angles. The horizontal pixel coordinate $u$ is defined as

\begin{equation}
    u = \frac{1}{2}\left(1-\frac{\operatorname{atan2}(y,x)}{\pi}\right)(w-1)
\end{equation}

where $w$ denotes the panorama pixel width. The vertical coordinate $v$ is computed from the elevation angle

\begin{equation}
    \phi = \arcsin\left(\frac{z}{r}\right)
\end{equation}

and mapped to the image grid as

\begin{equation}
    v = \left(1-\frac{\phi+f_{\mathrm{down}}}{f_{\mathrm{up}}+f_{\mathrm{down}}}\right)(h-1),
\end{equation}

where h is the panorama height and $f_{up} = f_{down} = 90^{\circ}$. This results in an equirectangular projection with a horizontal field of view of $360^{\circ}$ and a vertical field of view of $180^{\circ}$. In our implementation, the image resolution is controlled by an upscaling factor and a fixed aspect ratio. Because the point cloud is sparse, naturally, not all pixels are populated. The upscaling factor can be used to manually control the ratio of empty pixels.

The same projection pipeline is used to render multiple modalities from the point cloud, including color, intensity, depth, surface normals, semantic labels, and instance labels. For visualization, scalar attributes, such as intensity and depth, can be normalized over valid pixels and stored as grayscale images. 
Surface normals are linearly mapped from $\left[ -1,1 \right]$ to RGB values. Semantic and instance annotations are rendered as both colorized visualizations and machine-readable label-ID images.

For each point $ p = (x,y,z)$, the radial distance to the sensor is computed as the Euclidean distance. 
If multiple points fall within the same pixel, the visibility is resolved by depth ordering. Points are sorted by radial distance and rasterized from far to near, such that the closest point overwrites all others at the same image location. This produces a z-buffer-like rendering \cite{nischwitz_Computergrafik_2019} that is consistent with the scanner's viewpoint. Pixels without a projected point remain invalid. 
This procedure establishes a direct correspondence between the 3D point cloud and its 2D panoramic representation, enabling multimodal use of geometry, appearance, and annotations within a common dataset.

In addition to the most common equidistant projection, this dataset also supports a cube map representation of the 3D space, projected into a collection of six square image faces with significantly reduced distortion, which may be preferred for 2D scene analysis \cite{inuganti_JOPP3D_2026, kirillov_Segment_2023, radford_Learning_2021}. Scripts for reproducibly generating both projection types from the released point-wise arrays are available in the Dataset Hub~\cite{rauch_Rohbau3D_2026}.

\subsection{Dataset Statistics}

Rohbau3D-Semantics comprises 504 terrestrial laser scan scenes from 14 construction sites, containing a total of 12,485,778,128 points and 20,786 annotated object instances. The dataset provides point-wise semantic labels for 17 foreground classes and one background class (None), together with class-wise instance annotations for object-level analysis.

Table \ref{tab:site_stats} summarizes the overall point cloud statistics for the unprocessed full-scale dataset. Across all 504 scenes, the global averages are 24,773,369 points and 41.24 annotated instances per scene. Additionally, each full-resolution scan is accompanied by an index file that enables reproducible subsampling to a capped version with at most 15 million points per scene, thereby increasing homogeneity and reducing memory usage in large-scale learning pipelines. This capped version is referred to as \textit{"15M"}.

\begin{table}[t]
  \caption{Dataset statistics by site.}
  \label{tab:site_stats}
  \centering
  \small
  \setlength{\tabcolsep}{3pt}
  \renewcommand{\arraystretch}{1.15}
  \begin{tabular}{lrrrr}
    \hline
    Site & \multicolumn{1}{c}{Points} & \multicolumn{1}{c}{Instances} &
    \begin{tabular}{c}mean \\ Pts./scene\end{tabular} &
    \begin{tabular}{r}mean \\ Inst./scene\end{tabular} \\
    \hline
    site\_00 &   629,986,448 &  1,227 & 17,499,620 &  34.08 \\
    site\_01 &   224,578,587 &   562 & 10,694,220 &  26.76 \\
    site\_02 &   118,823,353 &   622 & 10,802,120 &  56.55 \\
    site\_03 & 1,549,189,153 &  3,366 & 27,664,090 &  60.11 \\
    site\_04 &   378,528,127 &   998 & 27,037,720 &  71.29 \\
    site\_05 & 1,771,487,032 &  2,134 & 27,253,650 &  32.83 \\
    site\_06 & 1,804,683,438 &  3,452 & 27,764,360 &  53.11 \\
    site\_07 &   135,670,833 &   513 & 27,134,170 & 102.60 \\
    site\_08 & 1,664,353,250 &  1,963 & 27,739,220 &  32.72 \\
    site\_09 & 1,009,946,470 &  1,064 & 28,054,070 &  29.56 \\
    site\_10 &   151,315,713 &   221 & 25,219,290 &  36.83 \\
    site\_11 & 1,122,217,527 &  1,242 & 26,098,080 &  28.88 \\
    site\_12 & 1,111,902,857 &  1,575 & 27,797,570 &  39.38 \\
    site\_13 &   813,095,340 &  1,847 & 17,675,990 &  40.15 \\
    \hline
    Tot./Avg. & 12,485,778,128 & 20,786 & 24,773,369 & 41.24 \\
    \hline
  \end{tabular}
\end{table}

\begin{figure*}[t]
    \centering
    \includegraphics[width=1.0\linewidth]{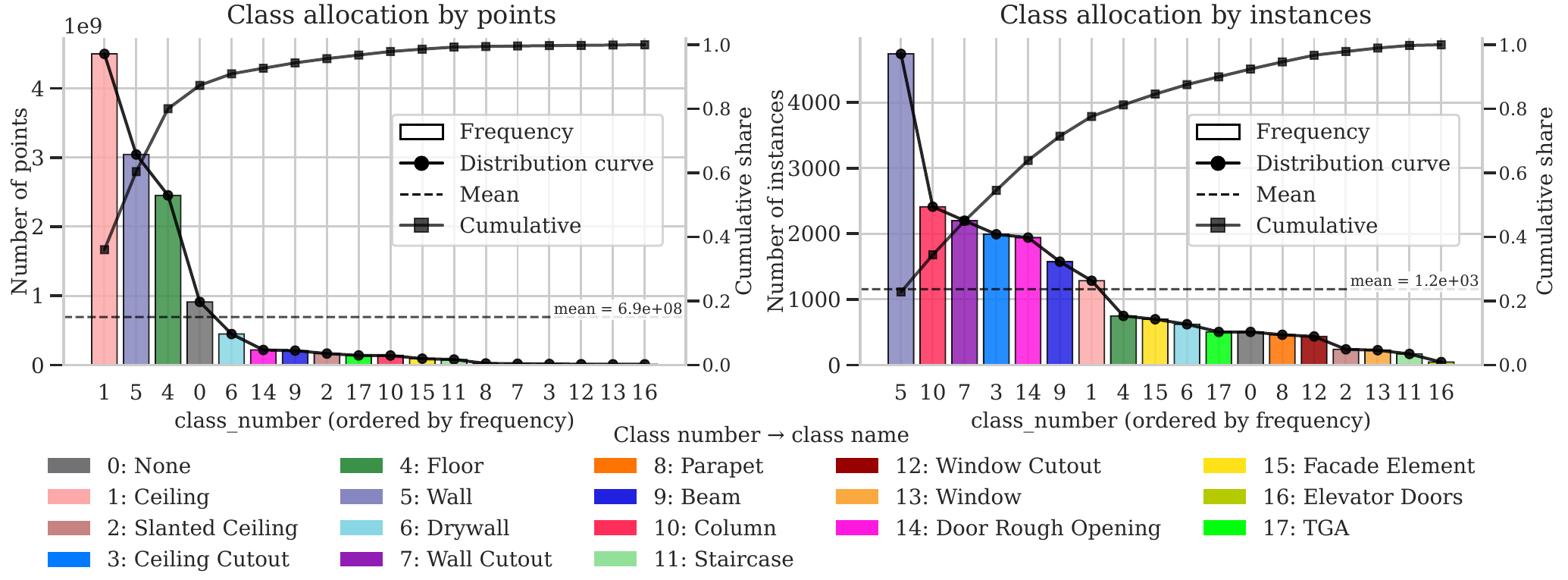}
    \caption{The allocation of all 18 object classes in terms of point- and instance-frequency recurrence in the Rohbau3D-Semantics data.}
    \label{fig:class_share}
\end{figure*}

Figure \ref{fig:class_share} illustrates the semantic class distribution in terms of both points and instances. The class distribution is skewed at the point level, with floor-, wall-, and ceiling-related classes accounting for the majority (80\%) of points.
The inherent nature of indoor datasets implies that most data points belong to one of these three classes, since these components constitute the vast majority of the convex building envelope. Rohbau3D is no exception; it is subject to the same imbalance as comparable indoor data sources. 
However, the instance-level distribution is substantially less skewed, which indicates that many smaller object categories recur frequently across the dataset despite contributing comparatively few points per instance.

The \textit{None} class plays a special role because all scene objects that cannot be assigned to any of the existing component classes are assigned to this \textit{None} class, combining not only all scan artifacts, ghost points, and measurement noise, but also all objects that are irrelevant to technically understanding the structure of the building. Overall, the \textit{None} class accounts for 7.3\% of the total points, although this percentage exceeds 10\% at some construction sites.

\section{Training, Validation \& Testing split}
\label{sec:split}

The 504 scans are not statistically independent: scans from different positions at the same construction site may capture overlapping areas and share distinctive geometry, materials, and construction conditions. A random scene-level split could therefore leak site-specific information and yield overly optimistic generalization estimates. Instead, we treat construction sites as independent units for testing and reserve all 96 test samples from site\_00 and site\_08 only for that. These two residential and high-rise office building projects were selected after examining the per-site instance distributions to obtain broad class coverage while retaining sufficient training data. Because no site contains every class and the number of scenes differs substantially among sites, the split prioritizes site-level independence and class coverage over equal scene counts or class frequencies.

The resulting allocation for our proposed standard split is 354/54/96 scenes (70.2\%/10.7\%/19.0\%) for training, validation, and testing. The following list defines the proposed split; numbers in parentheses indicate the number of validation scenes selected from each training site.
The full list of validation scene IDs is provided in \texttt{metadata/data\_split.json} in the Dataset Hub~\cite{rauch_Rohbau3D_2026}.

\begin{itemize}
  \setlength{\parskip}{0pt}
  \setlength{\itemsep}{0pt plus 1pt}
    \item \textbf{Train\,/\,Val (70\%\,+\,11\%):} site\_01 (3), site\_02 (2), site\_03 (7), site\_04 (1), site\_05 (9), site\_06 (10), site\_07 (1), site\_09 (4), site\_10 (1), site\_11 (6), site\_12 (5), site\_13 (5); \\ [-0.6em]
    \item \textbf{Test (19\%):} site\_00, site\_08. 
\end{itemize}

Validation is intended for model selection rather than as an independent estimate of cross-site generalization. Holding out further complete sites for validation would reduce the already limited pool of 14 sites and could remove rare classes from training or validation. We therefore selected 54 validation scenes from the 12 training sites using their class and instance distributions to retain coverage of the full taxonomy, leaving 354 scenes for training. Consequently, the training and validation datasets may share similar location features, even though we ruled out overlaps between neighboring scans, while the test dataset does not overlap with either subset by location or local features. Results on site\_00 and site\_08 should thus be interpreted as a reproducible fixed-split reference on two unseen projects, not as comprehensive evidence across all construction-site diversity. Site-level cross-validation over alternative held-out groups would provide stronger generalization evidence but is left for future work.

\section{Semantic Segmentation Baseline}

\begin{figure*}[t]
    \centering
    \includegraphics[width=\textwidth]{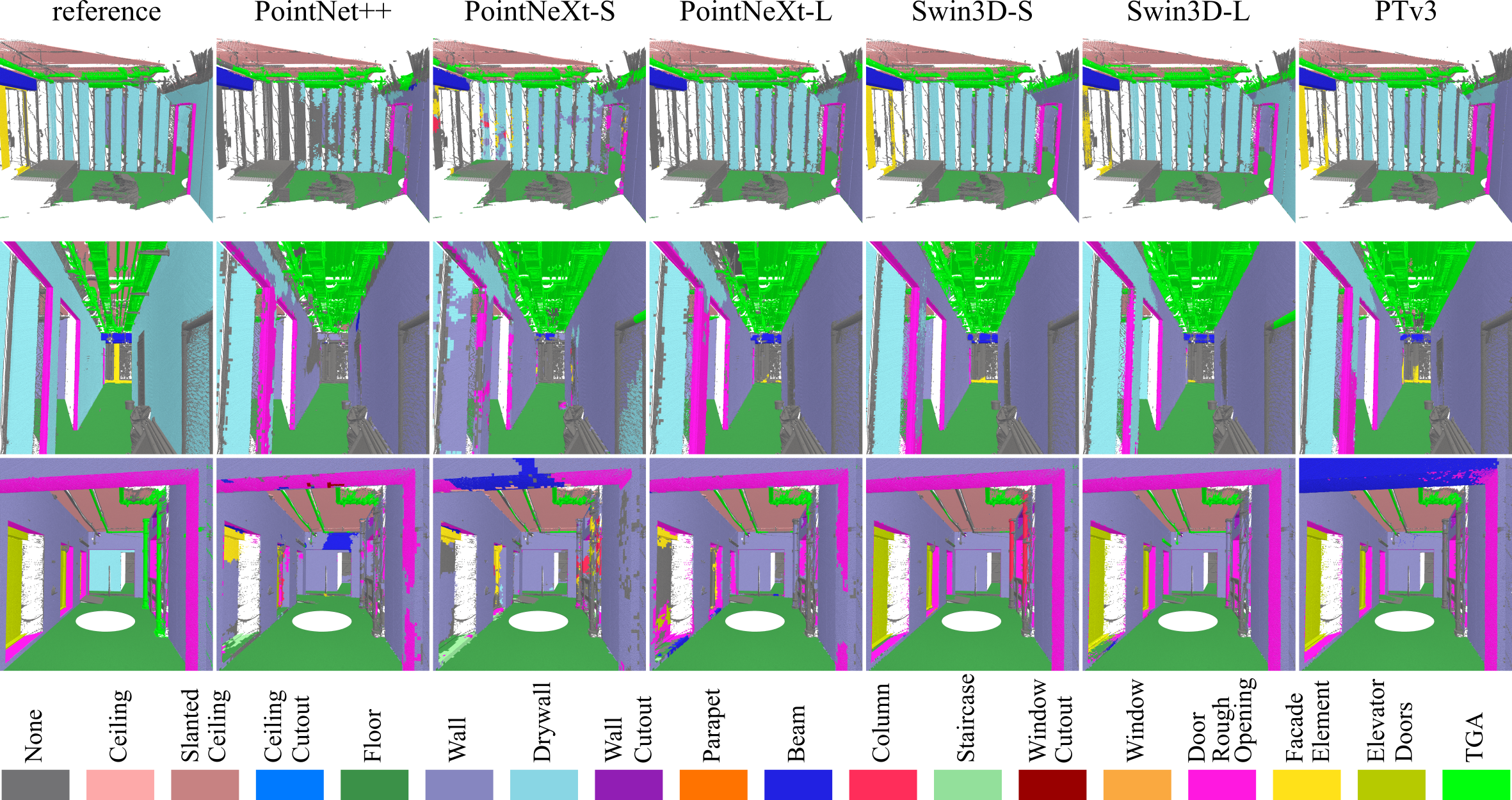}
    \caption{Qualitative baseline 3D-point cloud semantic segmentation results from supervised learning on the Rohbau3D-Semantics data.}
    \label{fig:results}
\end{figure*}

We report results for 3D point cloud semantic segmentation, serving as the initial baseline for this dataset, obtained from fully supervised training on 17+1 object classes. This baseline is evaluated on the capped 15M version of Rohbau3D-Semantics' proposed standard data split. It does not benchmark instance segmentation or isolate the contributions of intensity, normals, depth, and the projected 2D representations. To provide a computationally tractable and consistent comparison across all evaluated backbones, the inputs were voxelized in the data loader using a 4x4x4\,cm grid. This preprocessing kept training within the 32\,GB memory budget of the available Tesla V100-SXM3 GPUs; evaluation at the full native point density is outside the scope of this initial baseline.
The method followed a fully supervised training pipeline. The implementation leveraged the two established libraries for point cloud processing: PointNeXt \cite{qian.PointNeXt.2022} for MLP-based DNNs, and Pointcept (v.1.6.1)\cite{pointceptcontributors.Pointcept.2023} for state-of-the-art point-transformer models.  
We compare six backbones from the two libraries, including the standard PointNet++ \cite{qi_PointNet_2017} for reference, PointNeXt \cite{qian.PointNeXt.2022} in the small (S) and large (L) model configurations, with varying network depth scale, Swin3D \cite{yang.Swin3D.2023} in the small (S) and large (L) model configurations for a scaled window size of the sliding window mechanism, and the latest Point Transformer V3 \cite{wu.Point.2023} (PTv3) as the current benchmark for standard supervised learning. 
All training runs used geometric augmentation in the coordinate space and feature augmentation in the RGB color space, a cosine-scheduled learning rate, cross-entropy loss, and an AdamW optimizer. Table \ref{tab:baseline} tabulates the quantitative results of this baseline on the proposed dataset. Performance is calculated by the per-class intersection over union (IoU) metric. Figure \ref{fig:results} showcases qualitative results.

\begin{table*}[t]
  \caption{Quantitative results for supervised 3D semantic segmentation on the 96 test scans from the held-out sites site\_00 and site\_08.}
  \label{tab:baseline}
  \centering
  \setlength{\tabcolsep}{3.7pt} 
  \small
  \begin{tabular}{l|cccccccccccccccccc|c}
    \toprule
    \multicolumn{1}{l}{model}
    & \rotatebox{90}{None}
    & \rotatebox{90}{Ceiling}
    & \rotatebox{90}{Slanted Ceiling}
    & \rotatebox{90}{Ceiling Cutout}
    & \rotatebox{90}{Floor}
    & \rotatebox{90}{Wall}
    & \rotatebox{90}{Drywall}
    & \rotatebox{90}{Wall Cutout}
    & \rotatebox{90}{Parapet}
    & \rotatebox{90}{Beam}
    & \rotatebox{90}{Column}
    & \rotatebox{90}{Staircase}
    & \rotatebox{90}{Window Cutout}
    & \rotatebox{90}{Window}
    & \rotatebox{90}{Door Rough Open.}
    & \rotatebox{90}{Facade Element}
    & \rotatebox{90}{Elevator Doors}
    & \multicolumn{1}{c}{\rotatebox{90}{TGA}}
    & mIoU \\
    \midrule
    PointNet++  & 0.58 & 0.95 & 0.45 & 0.42 & 0.95 & 0.71 & 0.28 & 0.19 & 0.30 & 0.48 & 0.49 & 0.69 & 0.11 & 0.13 & 0.32 & 0.18 & 0.14 & 0.57 & 0.44
    \\
    PointNeXt-S & 0.61 & 0.95 & 0.48 & 0.32 & 0.95 & 0.76 & 0.46 & 0.11 & 0.44 & 0.47 & 0.52 & 0.64 & 0.06 & 0.06 & 0.29 & 0.23 & 0.02 & 0.57 & 0.44
    \\
    PointNeXt-L & 0.67 & 0.96 & 0.58 & 0.63 & 0.97 & 0.85 & 0.65 & 0.23 & 0.57 & 0.59 & 0.73 & 0.81 & 0.22 & 0.12 & 0.50 & 0.36 & 0.03 & 0.62 & 0.56
    \\
    Swin3D-S    & 0.75 & 0.98 & 0.73 & \textbf{0.83} & 0.97 & 0.87 & 0.68 & 0.37 & \textbf{0.76} & 0.64 & 0.81 & \textbf{0.93} & 0.43 & 0.32 & 0.60 & 0.59 & 0.45 & \textbf{0.69} & 0.69
    \\
    Swin3D-L    & 0.77 & 0.97 & 0.72 & 0.75 & 0.97 & \textbf{0.91} & \textbf{0.78} & 0.30 & 0.68 & \textbf{0.69} & 0.78 & 0.93 & 0.45 & \textbf{0.56} & \textbf{0.63} & 0.59 & 0.46 & 0.66 & 0.70
    \\
    PTv3       & \textbf{0.77} & \textbf{0.98} & \textbf{0.73} & 0.80 & \textbf{0.98} & 0.89 & 0.72 & \textbf{0.46} & 0.62 & 0.67 & \textbf{0.82} & 0.92 & 0.48 & 0.43 & 0.63 & \textbf{0.64} & \textbf{0.71} & 0.68 & \textbf{0.72}
    \\
    \bottomrule
  \end{tabular}
\end{table*}

The evaluated baselines yield mIoU values between 0.44 and 0.72, with PTv3 achieving the highest value. Across all models, large and structurally dominant classes such as \textit{Ceiling}, \textit{Floor}, and \textit{Wall} achieve consistently high IoU values, whereas smaller and semantically more specific categories, including \textit{Wall Cutout}, \textit{Window Cutout}, \textit{Window}, and \textit{Elevator Doors}, remain considerably more challenging. Given the pronounced class imbalance and the relative rarity of these categories, this pattern is expected. The results provide an initial quantitative reference for semantic segmentation difficulty across coarse and fine-grained classes under the proposed split; they do not establish performance for the dataset's instance-level or multimodal use cases.

\section{Conclusion}
\label{sec:conclusion}

We presented a dataset of small- to large-scale indoor spaces from construction-site environments during the structural work phase, with high-detail semantic class and instance annotations in both 3D-space and projected 2D representations. 
In contrast to existing indoor 3D datasets, these data preserve the characteristic properties of physical laser sensing, including irregular sampling, reflections, and measurement artifacts. At the same time, terrestrial laser scanning provides exceptionally high point density and geometric accuracy, resulting in a benchmark that is realistic, precise, and challenging. 
The annotations required 3,818 hours of professional manual labeling and were supported by expert supervision, a second review of all 504 scenes, and automated completeness and consistency checks. Because the full dataset was annotated by one provider and subsequently reviewed and corrected by the authors, we do not report inter-annotator agreement or accuracy against an independently labeled subset; annotation quality is instead supported by full-scene expert review and automated consistency checks. The dataset's scale further distinguishes it from previous construction-site datasets: it comprises 504 scans from 14 sites across multiple building types and construction settings.

Beyond extending the raw data with semantic and instance labels, we documented the acquisition pipeline, preprocessing steps, annotation workflow, panoramic projection procedure, and benchmark splits, and reported baseline results for supervised 3D point cloud semantic segmentation. The final dataset is relevant not only for construction-site analytics, digital documentation, and BIM-related workflows, but also for robust 3D scene understanding in cluttered, partially occluded, and geometrically incomplete real-world environments with challenging lighting conditions and sensor noise. 

The present evaluation is limited to supervised 3D semantic segmentation on one reference split. It therefore neither establishes an instance-segmentation benchmark nor quantifies the contributions of the available intensity, depth, normals, and projected 2D representations. The corresponding annotations and modalities are released as resources for these future evaluations, rather than presented as empirically validated contributions of the current baseline. Future work should establish instance-level baselines, modality ablations, and site-level cross-validation over alternative held-out groups. Recent developments in foundation models, promptable 3D segmentation, open-vocabulary 3D scene understanding, and self-supervised learning provide further opportunities to use the dataset beyond the present supervised setting.

Finally, although the dataset is based on stationary laser scanning rather than mobile LiDAR, it may also serve as a useful reference for studying transfer to lower-density robotic sensing scenarios. We hope that Rohbau3D-Semantics contributes to more realistic, application-relevant, and data-efficient methods for understanding complex built spaces.

\section{Licensing and Access}
\label{sec:license}

Rohbau3D-Semantics is released under the Creative Commons Attribution 4.0 International (CC BY 4.0) license, permitting reuse with attribution. The data are hosted in the university's institutional repository (\href{https://doi.org/10.60776/ZWJFI4}{doi:10.60776/ZWJFI4}); download and processing scripts are available on GitHub (\href{https://github.com/RauchLukas/rohbau3d}{github.com/RauchLukas/rohbau3d}).

\clearpage
{
    \small
    \bibliographystyle{ieeenat_fullname}
    \bibliography{references}
}

\clearpage
\appendix

\setcounter{figure}{0}
\renewcommand{\thefigure}{S\arabic{figure}}

\section{Construction-Site Overview}
\label{sec:supp_site_overview}

Table~\ref{tab:supp_site_description} summarizes the site characteristics most relevant to semantic diversity and LiDAR sensing difficulty. The complete machine-readable site metadata remain available through the accompanying Dataset Hub~\cite{rauch_Rohbau3D_2026}.

\setcounter{table}{0}
\renewcommand{\thetable}{S\arabic{table}}
\begin{table*}[!t]
  \caption{Site-level environmental diversity in Rohbau3D-Semantics. An asterisk marks a renovation site.}
  \label{tab:supp_site_description}
  \centering
  \footnotesize
  \setlength{\tabcolsep}{4pt}
  \renewcommand{\arraystretch}{1.12}
  \begin{tabular}{@{}>{\raggedright\arraybackslash}p{1.15cm}>{\raggedright\arraybackslash}p{5.0cm}>{\raggedright\arraybackslash}p{10.15cm}@{}}
    \toprule
    \textbf{Site} & \textbf{Building and construction} & \textbf{Distinctive geometry and sensing conditions} \\
    \midrule
    site\_00 & Multi-story apartment building; brick masonry & Small--medium rooms; partly plastered walls; windows installed, doors absent. \\
    site\_01 & Multi-story apartment building; brick masonry & Small--medium rooms; slanted ceilings; partly plastered walls; windows installed, doors absent. \\
    site\_02 & Underground parking structure; reinforced concrete & Column grid; variable ceiling heights; low light; floor puddles. \\
    site\_03 & Multi-story school; reinforced-concrete frame & Large rooms; well-lit areas; floor puddles. \\
    site\_04 & Indoor swimming-pool hall; reinforced concrete & Large open volume; rounded ceiling components; large floor opening; facade not installed. \\
    site\_05 & Multi-story school; drywall partitions & Variable room sizes; semi-transparent temporary facade covering. \\
    site\_06 & Multi-story school; mixed solid and drywall partitions & Medium--large rooms; open facade areas; ceiling-mounted technical equipment. \\
    site\_07 & Large sports hall; reinforced concrete & Large open volume and high ceiling. \\
    site\_08 & High-rise office building; drywall construction & Small--large rooms; freestanding drywall components; glazed facade and ceiling services installed. \\
    site\_09* & Historic multi-story masonry building & Small rooms; narrow staircases; windows installed, doors absent; low light. \\
    site\_10* & Vaulted brick cellar & Small rooms; uneven floors; low light. \\
    site\_11 & Two-story building with basement; masonry and precast concrete & Small--medium rooms; floor puddles; low light. \\
    site\_12 & Multi-story apartment building with basement; precast reinforced concrete & Large window and door rough openings; some floor puddles and low-light scenes. \\
    site\_13* & Multi-story masonry building & Small rooms connected by corridors; partly plastered and partly exposed walls. \\
    \bottomrule
  \end{tabular}
\end{table*}

\section{Dataset Structure and Format}
\label{sec:supp_dataset_structure}

The released data follow a site--scan hierarchy, illustrated below in abbreviated form. Each scan is an independent, scanner-centered observation: its origin is the scanner position, and scans are not supplied in a common registered site coordinate system. Consequently, instance identifiers describe component observations within one scan and must not be interpreted as persistent object identities across scans. The repository-level JSON files provide the class registry, benchmark split, point-cloud file inventory, and site metadata, while the site PDF files provide rendered feature overviews.

\par\vspace{\baselineskip}
\noindent\begin{minipage}{\columnwidth}
\footnotesize
\begin{verbatim}
rohbau3d/
|-- metadata/
|   |-- class_registry.json
|   |-- data_split.json
|   |-- pointcloud_file_list.json
|   |-- site_metadata.json
|   '-- panorama/features/
|       '-- site_<id>.panorama.features.pdf
|-- site_00/
|   |-- scene_00000/
|   |   |-- coord.npy, color.npy
|   |   |-- intensity.npy, normal.npy
|   |   |-- class.npy, instance.npy
|   |   |-- sample_idx.npy
|   |   |-- inv_sample_idx.npy
|   |   |-- panorama/ (*)
|   |   '-- cube_map/  (*)
|   '-- ...
|-- site_01/ ...
'-- site_13/
\end{verbatim}
\end{minipage}
\par\vspace{\baselineskip}

Within each scan directory, the feature and label arrays use a common point ordering: row $i$ in each point-wise array describes the same measured point. Table~\ref{tab:supp_file_format} summarizes the released NumPy format. A redundant scene-level JSON record containing the annotation metadata for every instance can also be selected by the repository downloader.

\begin{center}
  \captionsetup{hypcap=false}
  \captionof{table}{Per-scan point-cloud files. $N$ denotes the number of points in the full-resolution scan.}
  \label{tab:supp_file_format}
  \centering
  \scriptsize
  \setlength{\tabcolsep}{3pt}
  \renewcommand{\arraystretch}{1.08}
  \begin{tabular}{@{}p{2.55cm}p{1.20cm}p{4.05cm}@{}}
    \toprule
    \textbf{File} & \textbf{Entries} & \textbf{Stored information} \\
    \midrule
    \texttt{coord.npy} & $N \times 3$ & Scanner-centered Cartesian coordinates $(x,y,z)$. \\
    \texttt{color.npy} & $N \times 3$ & Point-wise RGB color. \\
    \texttt{intensity.npy} & $N$ scalar & Laser reflection intensity. \\
    \texttt{normal.npy} & $N \times 3$ & Reconstructed surface-normal vectors. \\
    \texttt{class.npy} & $N$ integer & Semantic label ID: 17 foreground classes and \textit{None}. \\
    \texttt{instance.npy} & $N$ integer & Scene-local instance ID; not a cross-scan object identity. \\
    \texttt{sample\_idx.npy} & indices & Reproducible selection of at most 15 million full-resolution points. \\
    \texttt{inv\_sample\_idx.npy} & indices & Index mapping from the 15M representation to the full-resolution point set. \\
    \bottomrule
  \end{tabular}
\end{center}

The full-resolution arrays remain the primary data; the two index files provide an optional computationally tractable 15M representation while retaining correspondence to the original observations.

The directories marked (*) are derived representations generated locally from the 3D arrays and are not separate downloadable measurements. The supplied rendering script can produce color, radial depth, intensity, normal, semantic-label, and instance-label panoramas or six-face cube maps. Panoramas default to $4096 \times 2048$ pixels and cube-map faces to $1024 \times 1024$ pixels; empty pixels are stored as transparent. The panoramic output additionally distinguishes visualized label images (\texttt{class.png}, \texttt{instance.png}) from machine-readable label-ID images (\texttt{class\_id.png}, \texttt{instance\_id.png}), while \texttt{depth.npy} retains numeric depth. Because all of these views are rendered from the scanner-centered point cloud using a common projection, they remain geometrically linked to the native 3D observations rather than constituting independent modalities.

\section{Annotation Review and Quality Control}

Quality control proceeded as an iterative review--correction cycle. After provider delivery, the authors inspected each scene in 3D using its semantic and instance views. Review targeted missing labels, semantic inconsistencies across scenes, incorrect instance merges or splits, boundary leakage, and isolated outliers that could distort an instance bounding box. Minor errors were corrected directly by the authors; scenes with broader or repeated problems were rejected with visual feedback and returned to the provider for correction. Corrected scenes were reviewed again before acceptance, and ambiguous construction elements were resolved with domain knowledge and incorporated into subsequent annotation decisions.

\begin{figure*}[p]
  \centering
  \includegraphics[height=0.895\textheight]{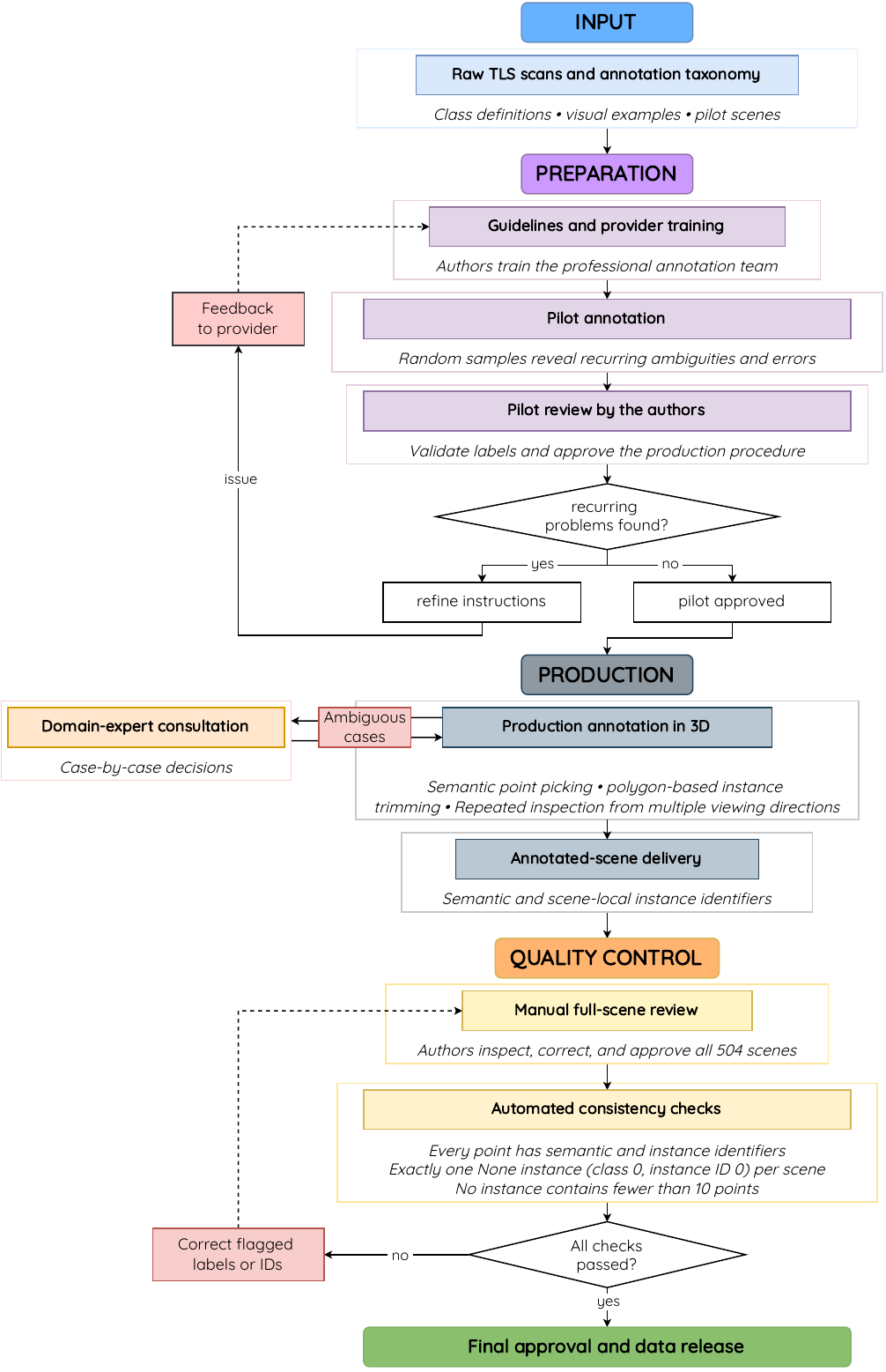}
  \caption{Annotation and quality-control pipeline. The workflow begins with taxonomy definition, annotation guidelines, provider training, and pilot scenes. Pilot review and construction-domain decisions provide feedback during annotation. After delivery, all 504 scenes undergo manual review, correction where necessary, and automated checks for complete semantic and instance identifiers, exactly one None instance per scene, and the absence of instances containing fewer than ten points. Failed checks trigger another correction and verification cycle before final approval.}
  \label{fig:supp_annotation_pipeline}
\end{figure*}

\begin{figure*}[t]
  \centering
  \includegraphics[width=\textwidth]{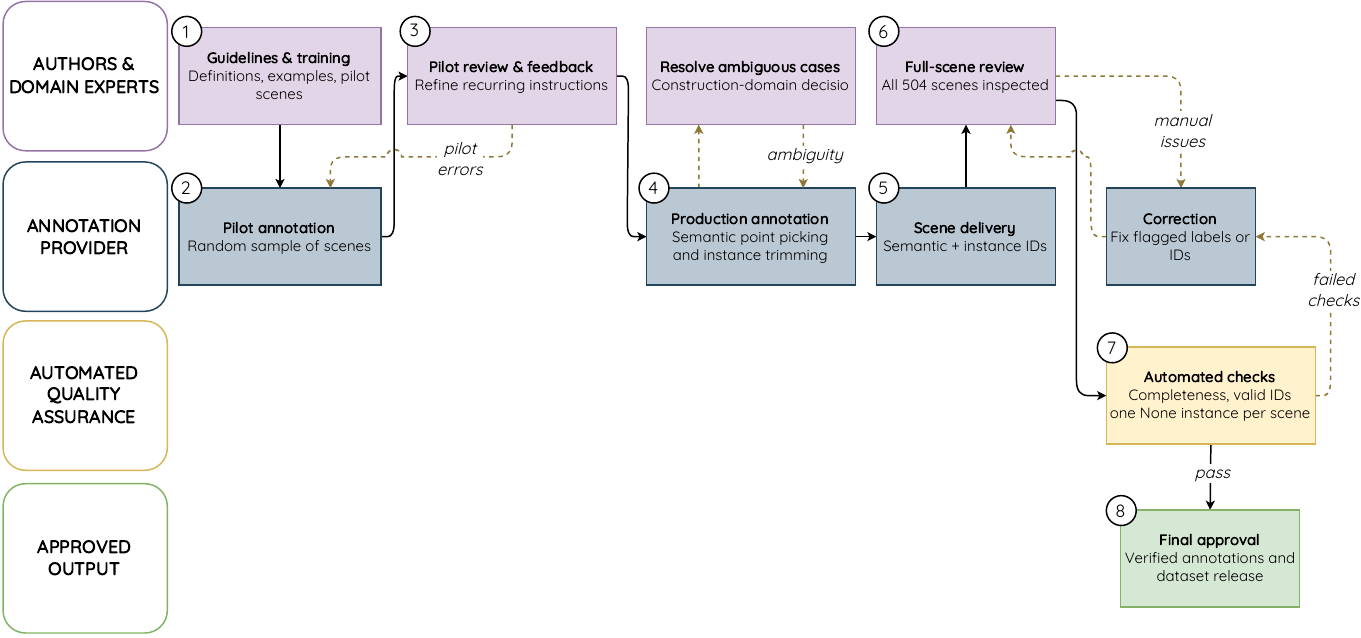}
  \caption{Responsibilities in the annotation and quality-control workflow. The swimlanes distinguish tasks performed by the authors and construction-domain experts, the professional annotation provider, and the automated quality-assurance procedure. Solid arrows show the primary workflow, while dashed arrows show feedback and correction loops initiated by pilot errors, ambiguous cases, manual review, or failed automated checks.}
  \label{fig:supp_annotation_responsibilities}
\end{figure*}

After manual acceptance, automated checks verified that every point had valid semantic and instance identifiers, each scan contained exactly one \textit{None} instance with ID~0, and no residual instance contained fewer than ten points. A failed check returned the scene to the correction loop. Figures~\ref{fig:supp_annotation_pipeline} and~\ref{fig:supp_annotation_responsibilities} summarize the sequence and division of responsibilities.

\section{Condensed Annotation Guidelines}

This section condenses the finalized instructions used to guide the professional annotation provider. The task was to assign every measured point in each raw terrestrial laser scan to one of the 17 foreground classes or the \textit{None} background class and to form scene-local object instances. The foreground taxonomy targets installed shell-construction components and technical building equipment; its complete class order and definitions are given below.

\subsection{General Annotation Principles}

\begin{enumerate}[leftmargin=*,itemsep=2pt,topsep=3pt]
  \item \textbf{Preserve the raw observations.} Annotate each scan as delivered: do not filter, delete, move, or otherwise clean points. Outliers, reflections, and ghost geometry remain in the data and are labeled; occlusions and incomplete regions are not reconstructed.

  \item \textbf{Prioritize precise boundaries.} Label large and small components at the point level and refine their boundaries as closely as the observations permit. Inspect geometry from multiple directions in 3D; use RGB appearance only as supporting evidence.

  \item \textbf{Segment objects, not only classes.} Distinguishable physical components receive separate instance identifiers even when they share a semantic class; for example, five columns are five instances. The exceptions are the single \textit{None} instance with ID~0 and the local grouping rule for related TGA elements described below.

  \item \textbf{Use \textit{None} rather than guessing.} Assign scan artifacts, stray points, objects outside the taxonomy, and geometry that cannot be classified reliably to \textit{None}. No point may remain unlabeled.

  \item \textbf{Restrict foreground labels to installed components.} Temporary supports, scaffolding, safety equipment, stored or not-yet-installed materials, loose cables, coverings, and comparable construction-site clutter belong to \textit{None}, not to the corresponding structural or TGA class.

  \item \textbf{Apply decisions consistently.} Use the same class and instance criteria across all scenes. Annotators were instructed to flag uncertain cases for review rather than resolve them by assumption; reviewed decisions were then applied to comparable cases.

  \item \textbf{Favor quality over throughput.} Accurate object boundaries and fine structures take precedence over annotation speed. Each segment should be checked before completion, particularly around occlusions, intersections, reflective surfaces, and small components.
\end{enumerate}

\subsection{Instance and Boundary Rules}

\paragraph{Instance formation.}
Instance identifiers are local to one scan and do not encode cross-scan object identity. A separate instance was created for each distinguishable physical component; TGA is the exception and uses local groups of related elements as described below. Touching elements remained separate when they represented different construction components, such as walls meeting at a corner or beams intersecting at a joint. A gap caused by occlusion or missing measurements split an instance only when the common physical connection was unclear or ambiguous; fragments were merged when their continuation as one component could be inferred reliably from the surrounding geometry. Each spatially distinguishable cutout or rough opening formed a separate instance. All \textit{None} points in a scan were collected in the single background instance with ID~0. Technical building equipment was grouped into local instances of related elements; close-by elements were combined when the scan did not support a reliable separation.

\paragraph{Cutout and rough-opening labels.}
These classes describe the narrow measured reveal or framing surfaces around an intentional opening---typically the two lateral jambs, the head or soffit, and the sill or lower reveal where present. They do not label empty space. If an otherwise empty opening was covered by temporary boards, foil, or comparable site material, points on that covering were assigned to \textit{None}, not to the cutout or rough-opening class.

\subsection{Class-Specific Rules}
The classes below follow the identifiers and order of the released class registry.

\begingroup
\small
\setlength{\parskip}{2pt}

\noindent\textbf{0 --- None.}
Background for sensor noise, isolated outliers, reflections and ghost geometry, temporary supports or safety equipment, stored and not-yet-installed materials, objects outside the taxonomy, and ambiguous points. All such points share instance ID~0 within a scan.

\noindent\textbf{1 --- Ceiling.}
Horizontal surfaces forming the upper interior boundary of a space. Beams, cutouts, technical equipment, and temporary shoring are excluded. Continuous observations belong to one instance unless a true geometric break or an ambiguous occlusion prevents a reliable connection.

\noindent\textbf{2 --- Slanted Ceiling.}
Inclined surfaces that function as the upper interior boundary, including surfaces beneath pitched roofs. Inclined beams and the underside of a staircase are excluded when their structural identity can be determined from context. Instance formation follows the ceiling rule.

\noindent\textbf{3 --- Ceiling Cutout.}
Measured reveal surfaces framing an intentional opening through a ceiling, such as a shaft, stair opening, or service penetration. The void and any temporary covering are excluded. Each distinguishable cutout is a separate instance.

\noindent\textbf{4 --- Floor.}
Horizontal, walkable structural surfaces. Rubble, tools, stored material, temporary coverings, and other clutter are excluded. A floor continuing through an open transition remains one instance; true gaps, stairs, or elevation changes establish a new instance.

\noindent\textbf{5 --- Wall.}
Solid vertical elements made of concrete, brick, masonry, or comparable construction. Surface channels, protrusions, and local imperfections remain part of the wall rather than being discarded for deviating from a plane. Each straight wall section is a separate instance, with corners or direction changes defining component boundaries.

\noindent\textbf{6 --- Drywall.}
Non-load-bearing gypsum-board partitions, including unfinished partitions with visible stud framing. Studs belonging to an incomplete drywall assembly are assigned to the adjoining drywall. Instance boundaries follow the wall rule, and RGB appearance may support the distinction from concrete or masonry.

\noindent\textbf{7 --- Wall Cutout.}
Reveal surfaces framing an intentional technical opening through a wall that is not intended for a window or door. Each spatially separate opening is an individual instance; empty space and temporary coverings are excluded.

\noindent\textbf{8 --- Parapet.}
Low, wall-like structural elements that do not extend to the ceiling, including protective barriers and low dividers. Full-height elements are labeled as walls. Each straight parapet section forms a separate instance, with corners starting a new instance.

\noindent\textbf{9 --- Beam.}
Permanent load-bearing elements that span between walls, columns, or other supports and exhibit a structural cross-section. Temporary supports and elongated non-structural objects are excluded. Distinct beams remain separate instances even where they meet or intersect.

\noindent\textbf{10 --- Column.}
Vertical load-bearing supports with a compact cross-section, including round, square, rectangular, freestanding, and partially wall-embedded columns. Every distinguishable column is a separate instance.

\noindent\textbf{11 --- Staircase.}
Permanent stair flights comprising treads, risers, and structurally associated landings. Temporary or permanent railings are excluded from this class. A connected flight and its associated landing may form one instance, whereas clearly separate flights form separate instances.

\noindent\textbf{12 --- Window Cutout.}
Reveal surfaces framing an opening intended for a window before installation. Each window location is a separate instance. The empty interior and temporary boards or foil covering the opening are excluded.

\noindent\textbf{13 --- Window.}
An installed or partially installed window unit, including its frame, sash, and glazing. Connected panes belonging to one assembled unit may share an instance; spatially distinct window units are separate. Reflection artifacts behind glazing are assigned to \textit{None}.

\noindent\textbf{14 --- Door Rough Opening.}
Reveal surfaces framing an opening intended for a conventional door before its installation. Oversized open passages that are not purpose-made door openings remain part of the adjacent wall classification. Each doorway is a separate instance, while its void and temporary covering are excluded.

\noindent\textbf{15 --- Facade Element.}
Installed near-full-height facade units, commonly composed of glazed, wooden, plastic, or metal panels. Walls, columns, beams, and parapets that bound the unit retain their respective classes. Each structurally unified facade unit forms an instance; reflection artifacts are assigned to \textit{None}.

\noindent\textbf{16 --- Elevator Doors.}
The complete installed elevator-doorway assembly, including the sliding door panels and metal framing. Each elevator entrance is a separate instance, while the surrounding wall or recess retains its applicable structural class.

\noindent\textbf{17 --- TGA.}
Installed technical building equipment that transports air, water, electricity, or data, including pipes, ventilation ducts, cable trays, and conduits. Related elements are organized into local instance groups, and close-by elements may share an instance when their individual boundaries are ambiguous. Loose cables, stored pipes, and temporary services are assigned to \textit{None}.

\endgroup

\begin{figure*}[p]
  \centering
  \includegraphics[width=\textwidth]{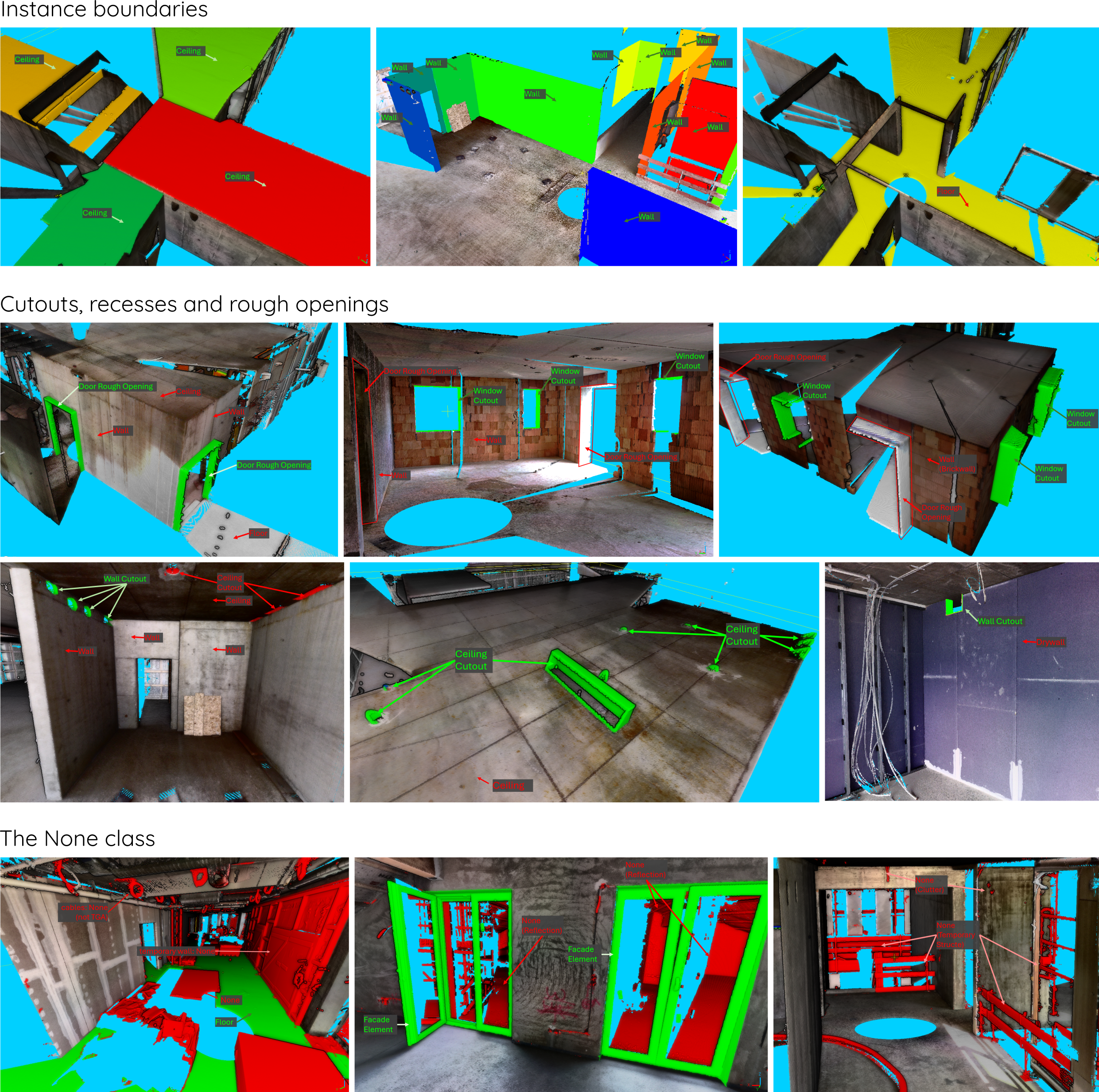}
  \caption{Representative cases from the finalized annotation guidelines. Top: different colors indicate scene-local instance identifiers; distinct ceiling and wall components are separated, whereas connected floor observations remain one instance. Middle: door rough openings and window, wall, and ceiling cutouts comprise the measured reveal surfaces, while surrounding structural components retain their respective classes. Bottom: installed facade elements receive the corresponding foreground label, whereas reflections, temporary structures, loose cables, clutter, and uninstalled components are assigned to \textit{None}. Arrows and labels identify the relevant semantic assignments.}
  \label{fig:supp_annotation_examples}
\end{figure*}

\hfill \break
Figure~\ref{fig:supp_annotation_examples} complements the class definitions with representative decisions concerning instance boundaries, cutout surfaces, installed components, and the \textit{None} class.

\end{document}